\documentclass[twocolumn,showpacs,preprintnumbers,amsmath,amssymb]{revtex4}

\usepackage{graphicx}
\usepackage{dcolumn}
\usepackage{bm}
\pdfoutput=1

\begin{document}

\preprint{APS/123-QED}

\title{ A Comparison of natural (english) and artificial (esperanto) languages. A Multifractal method  based analysis}

\author{J. Gillet}
\email{jgillet@ulg.ac.be}
\altaffiliation{Present address : Institut de Physique Nucl\'eaire, Atomique et de Spectroscopie, Universit\'e de Li\`ege, B-4000 Li\`ege, Belgium}

\author{M. Ausloos}
\email{Marcel.Ausloos@ulg.ac.be}

\affiliation{%
GRAPES,  B5a Sart-Tilman, B-4000 Li\`ege, Belgium
}%

\date{\today}

\begin{abstract}
We present a comparison of  two english texts, written by Lewis Carroll, one (Alice in wonderland) and the other (Through a looking glass), the former translated into esperanto, in order to observe whether natural and artificial languages  significantly differ from each other. We construct one dimensional time series like signals using either word lengths or word frequencies. We use the multifractal ideas for sorting out correlations in the writings. In order to check the robustness of the methods we also  $write$ (!) (\textit{consider?})
 the corresponding shuffled texts. We compare characteristic functions and e.g. observe marked differences in the (far from parabolic) $f(\alpha)$ curves, differences which we attribute to Tsallis non extensive statistical features in the  frequency time series and length time series.  The esperanto text has more extreme vallues. A very rough approximation consists in modeling the texts as a random Cantor set if resulting from a binomial cascade of long and short words (or words and blanks). This leads to parameters characterizing the text style, and most likely \textit{in fine} the author writings. 

\end{abstract}


\maketitle


As soon as modern fractals appeared in order to describe physical objects, it was evident that some generalization was in order:  multifractals spurred up, e.g., since obvioulsy a fractal dimension $D$ is not enough to describe an object \cite{halsey,west}. The more so in non equilibrium systems, characterized by some unusual dynamics.
Through a generator and from an initiator one can produce a fractal object with a given dimension.  How to produce realistic and meaningful multifractal models is a challenge.
Do they really exist \cite{luxma}? Do multifractal model exist nowadays \cite{lux2}? These questions come in parallel with the measurement of the fractal dimension, ... and its distribution. 
One question of interest is whether the apparently multifractal nature of an object is due to its
finite size or to a complex dynamical feature or something else!
Some attempt in this direction results from observation of multifractal features in meteorology and climate studies \cite{nadia, clouds,kava}, but also in many other fields \cite{bunde}, like mathematical finance \cite{lux2,makiCPC,oilprice,drozdz,canessa,KIMA}

Let us recall that one has basically to obtain a $D(q)$  function or  the $f(\alpha)$ spectrum, where $q$ represents the degree of some moment distribution of some variable, and $\alpha$  is some sort of critical exponent at phase transitions, also called the $Holder$ exponent; $f(\alpha)$  being its distribution.

There is a need for experimental work leading to reliable $D(q)$ and $f(\alpha)$ data, before modeling. Interesting  pioneering data should be here recalled : see work on DLA \cite{DLA1,DLA2}, DNA \cite {DNA,rosas}, SOI \cite{SOI}, NAO \cite{NAO}, .... It appears that most of the time some ``signal is either directly a time series or is transformed into a time series; more generally, the signal is called a text, because it can be decomposed through level thresholds which can be thought to be a set of characters taken from an alphabet. Here below we take real texts in fact as the source of experimental observations and follow the multifractal ideas to make an analysis of such texts. The main question concerns whether multifractals are indeed found in real texts; a question raised in \cite{ebeling2}; another is whether the technique of analysis can give some insight on a logical construction \cite{saakian}, from which stems the possible connection of such ideas with coding, transcription, machine translation, social distances,  network properties of languages, ... \cite{saakian} \cite{zlatic}  \cite{rodgers}  and more classically in physics about identifying coherent structures in spatially extended systems \cite{Shalizi}.   We examine two classical english texts but also one translation, into an unusual language, and the corresponding shuffled texts.  We focus on how local/structural properties develop into global ones.

Since Shannon \cite{shannon}, writings and codings are of interest in statistical physics.  Writings are systems practically composed of a large number of internal components (the words, signs, and blanks in printed texts). In terms of complexity investigations, writings which are a form of recorded languages, like living systems,  belong to the top level class involving highly optimized tolerance design, error thresholds in optimal coding, and  financial markets \cite{saakian}.  Relevant questions pertain to the life time, concentration, distribution, .. complexity of these.  One should distinguish two main frameworks. On one hand, language developments seem to be understandable through competitions, like in Ising models, and in self-organized systems. Their diffusion seems similar to percolation and  nucleation-growth problems  taking into account the existence of different time scales, for  inter- and intra- effects; this is the realm of anthropology. The second frame originates from  more classical linguistics studies; it pertains to the content and meanings of words and texts. Concerning the internal structure of a text, supposedly characterized by the language in which it is written,  it is well known that a text can be mapped into a signal, of course through the alphabet characters. However it can be  also reduced to less abundant symbols through some threshold, like a time series, which can be a list of +1 and -1, or sometimes 0.   
In fact,  laws of text content and structures have been searched for  a long time ago by Zipf  and others, see many refs. in \cite{JGMA1}, through the least effort (so called ranking) method. The technique is now currently applied in statistical physics as a first step to obtain, when they exist, the primary scaling law. Yet long range order correlations (LROC) between words in text are searched for. In \cite{amit} it was claimed that LROC express an author's ideas, and {\it in fine} consist in some author's signature.

Interestingly writings can be thought as social networks \cite{rodgers}. Social networks have fractal properties \cite{boguna}; most usually they should be multifractals; one can thus imagine/consider that a text which is a form of partially self-organized social network (for words) due to grammatical and style constraints present multifractal features. The properties of such texts taken as signals have already been examined,  e.g., a multifractal analysis of \textit{Moby Dick} letter distribution can be found  in \cite{ebeling2}.   
 
Even though we recognize such a pioneering paper, we stress that sentences made of words, not letters, are translated. Thus we present below an original consideration in this respect, i.e. the analysis and results about a translation between one of the most commonly used language, i.e. english, and a relatively recent language, i.e. esperanto. Esperanto is an artificially constructed language \cite{esperanto}, which was intended to be an easy-to-learn lingua franca. Statistical analyses seem to indicate that esperanto's statistical proportions are similar to those of other languages \cite{manaris}.  It was found that esperanto's statistical proportions resemble mostly those of German and Spanish, and somewhat surprisingly least those of French and Italian. English seems to be $the$ intermediary case. 
 
Comparison of different languages (writings) arising from apparently different origins or containing  different signs, e.g. greek \cite{greek}, turkish \cite{turkish}, chinese \cite{chinese}, ...  even somewhat artificial $languages$,    like those used for simulation codes on computers \cite{fortran} has also been made.
To our knowledge few comparisons have been presented about $written$ texts translated from one to another language \cite{footnote1,newIJMPC}  and in particular from the point of view of LROC in words. 

 The text used here was chosen for its wide diffusion, freely available from the web \cite{Gutenberg} and as a representative one of a famous scientist, Lewis Caroll, i.e. \textit{ Alice in wonderland} (AWL) \cite{carollAWL}. Moreover knowing the special (mathematical) quality of this author mind, and some, as we thought \textit{a priori}, some possibly special way of writing, another text has been chosen for comparison, i.e.  \textit{Through a looking glass} (TLG)  \cite{carollTLG}; - to our knowledge only available in english (on the web). This will allow us to discuss whether the differences, if any, between esperanto and english, are apparently due to the translation or to the specificity of this author work. 
 Previous work on  the english AWL version (AWL$_{\mbox{eng}}$) should be mentioned \cite{powers}, but pertains to a mere Zipf analysis.

 In Sect. \ref{sec-1}, we  present some elementary facts on these texts and  briefly expose the methodology, i.e. we  emphasize that we distinguish between ``frequency time series (FTS) and ``length time series (LTS). We  recall the multifractal technique for this specific application.  In order to check the robustness of the method we also invent  ($write$ !) (or \textit{consider}) the corresponding $shuffled$ texts, to which we apply the same technique of analysis.
 Therefore, in Sect. \ref{sec-2}, we present the somewhat unusual results, and discuss them in Sect. \ref{sec-3}. For the simplicity of the discussion we very roughly approximate the texts as resulting from a binomial cascade of short and long words. We obtain parameters {\it in fine}  characterizing the text style and the author's writings.
We observe that such multifractals have a deep connection to Tsallis non extensive statistics as pointed out  in ref. \cite{erzan} in another framework.

 \section{Data and Methodology} \label{sec-1}
 
 For our empirical considerations we have selected the two texts here above mentioned downloading them from a freely available site \cite{Gutenberg}, resulting obviously into three files, called. Next, we have removed the chapter heads. All our analyses are carried over this reorganized file  for each text.  Thereafter, we have shuffled these texts, in the files, without taking into account the punctuation \cite{footnote2}.
 
There  are two ways  to 
construct a time series from such documents  
\begin{enumerate}
 \item Take a document of $N$ words. Select all the different words. Count the frequency $f$ of appearance of each word in the document.   The $time$ of  ``appearance  is played by the rank position of the word in the file. We map the word frequency to a time series $f(t)$.   Such a time series is called the frequency time series (FTS).
\item Take a document of $N$ words. Consider the length $l$ (number of letters) of each word. Record $where$ each word of length $l$ is located in the text;  the $time$ is
played by the position of the word in the document, i.e. the first word is considered to be emitted at time $t $= 1, the second at time $t$ = 2, etc. A time series $l(t)$ is so
constructed. We refer to such a time series as the length time series (LTS).
 \end{enumerate}
 
\begin{figure}
\includegraphics[width=2.4in]{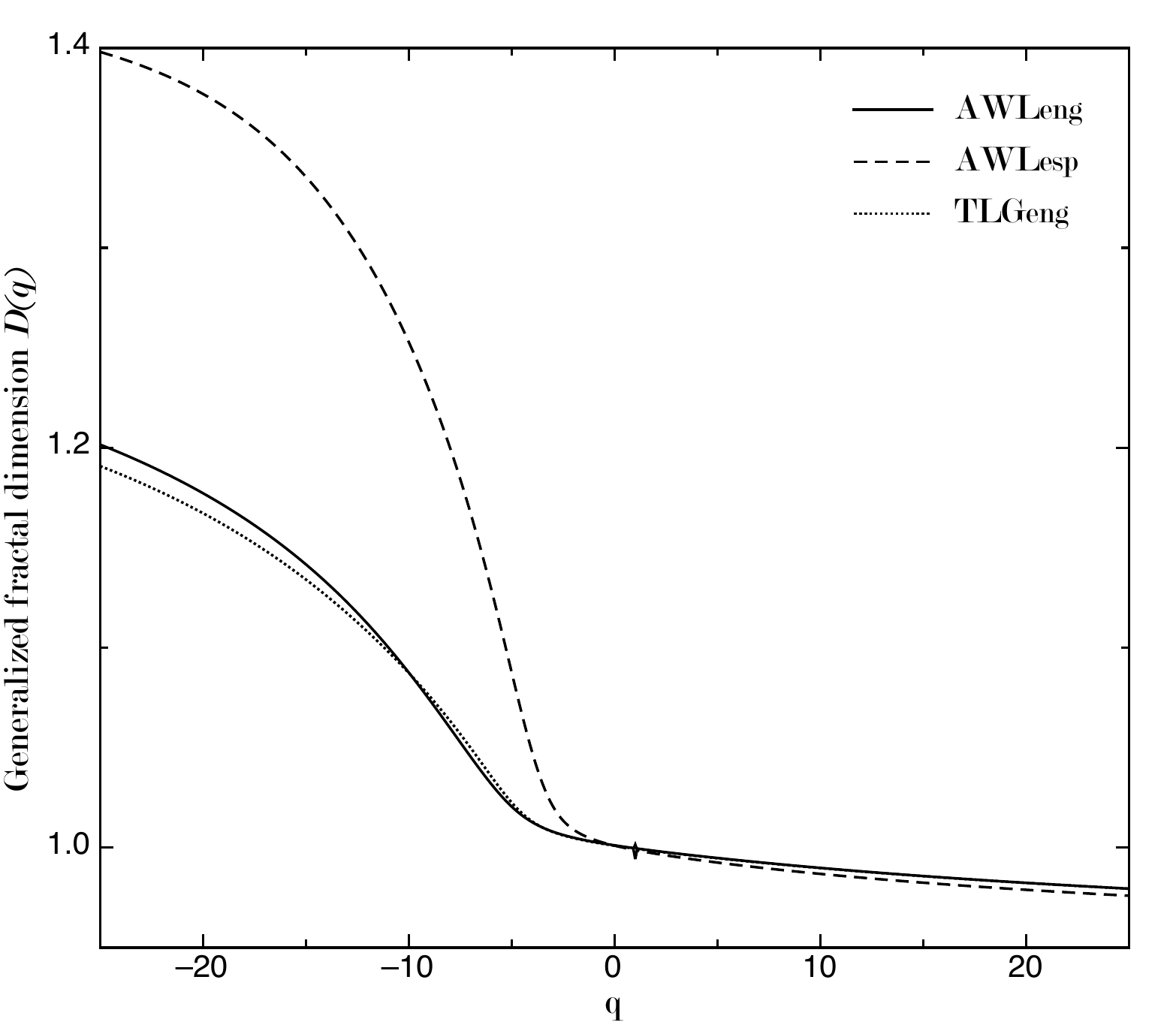}
\vspace{-0.4cm}
\includegraphics[width=2.4in] {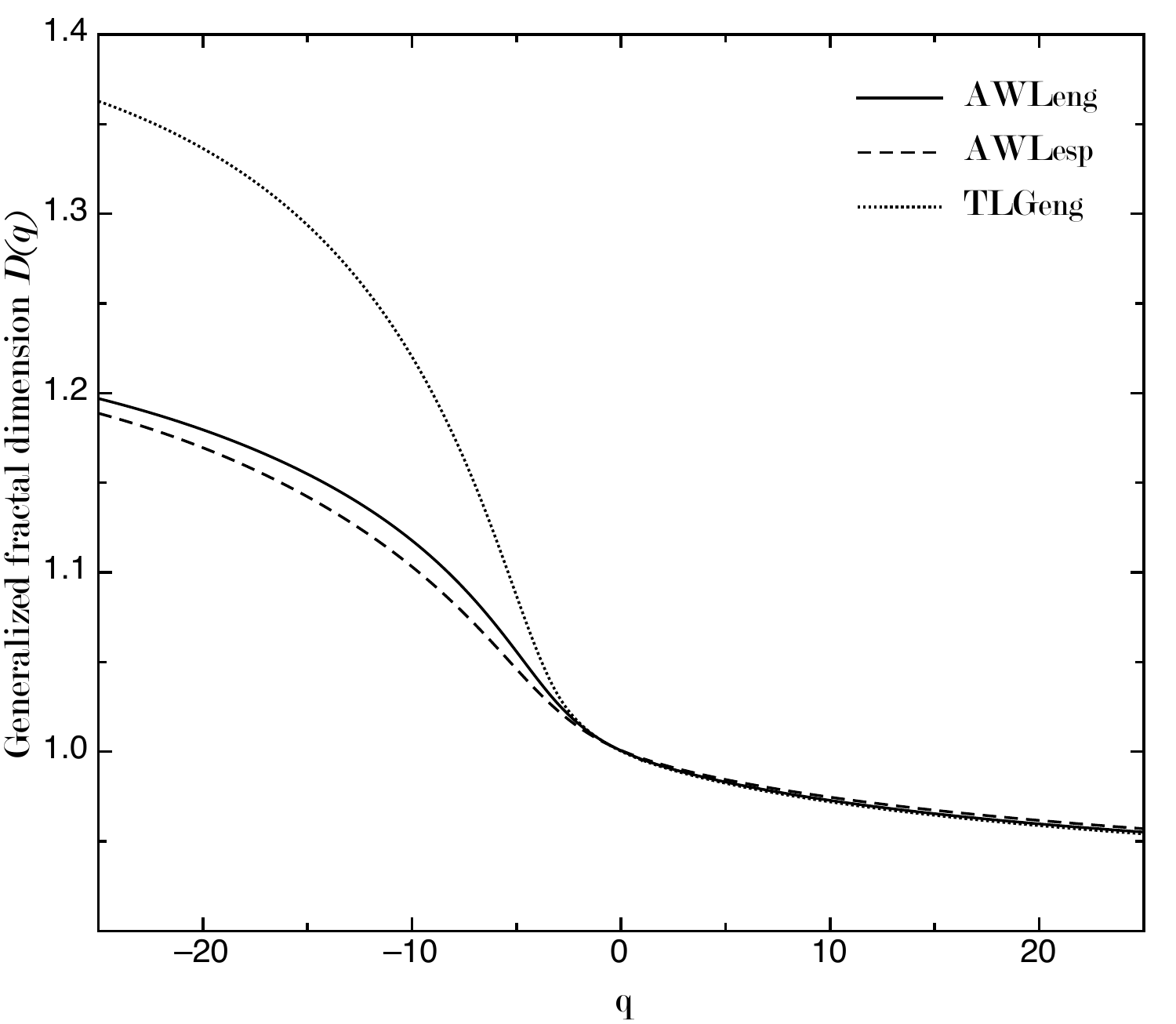}
\caption{\label{fig-1} D(q) for (a)  FTS (b) LTS  of the  original texts }
\end{figure}

\begin{figure}
\includegraphics[width=2.4in] {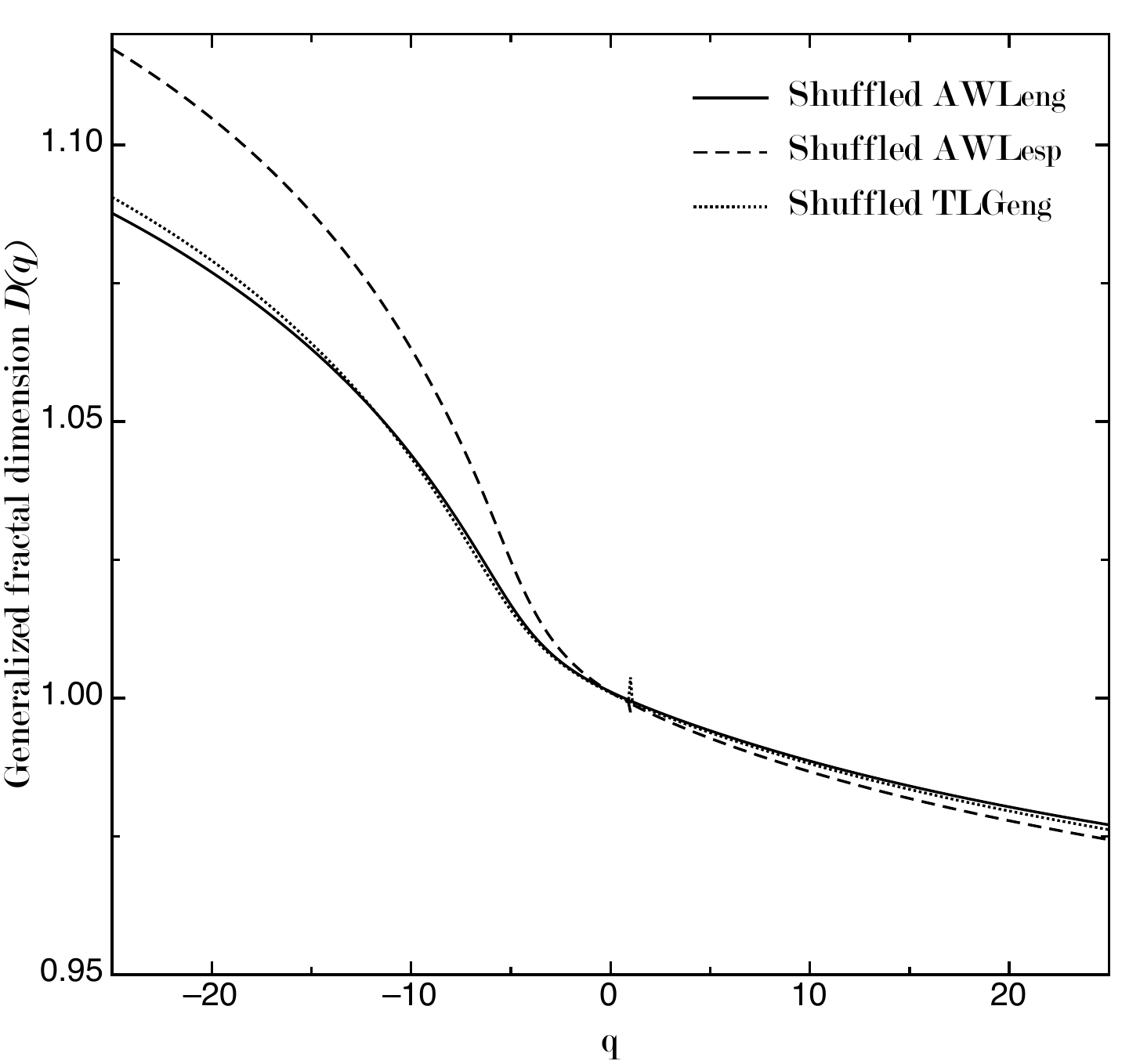}
\vspace{-0.4cm}
\includegraphics[width=2.4in] {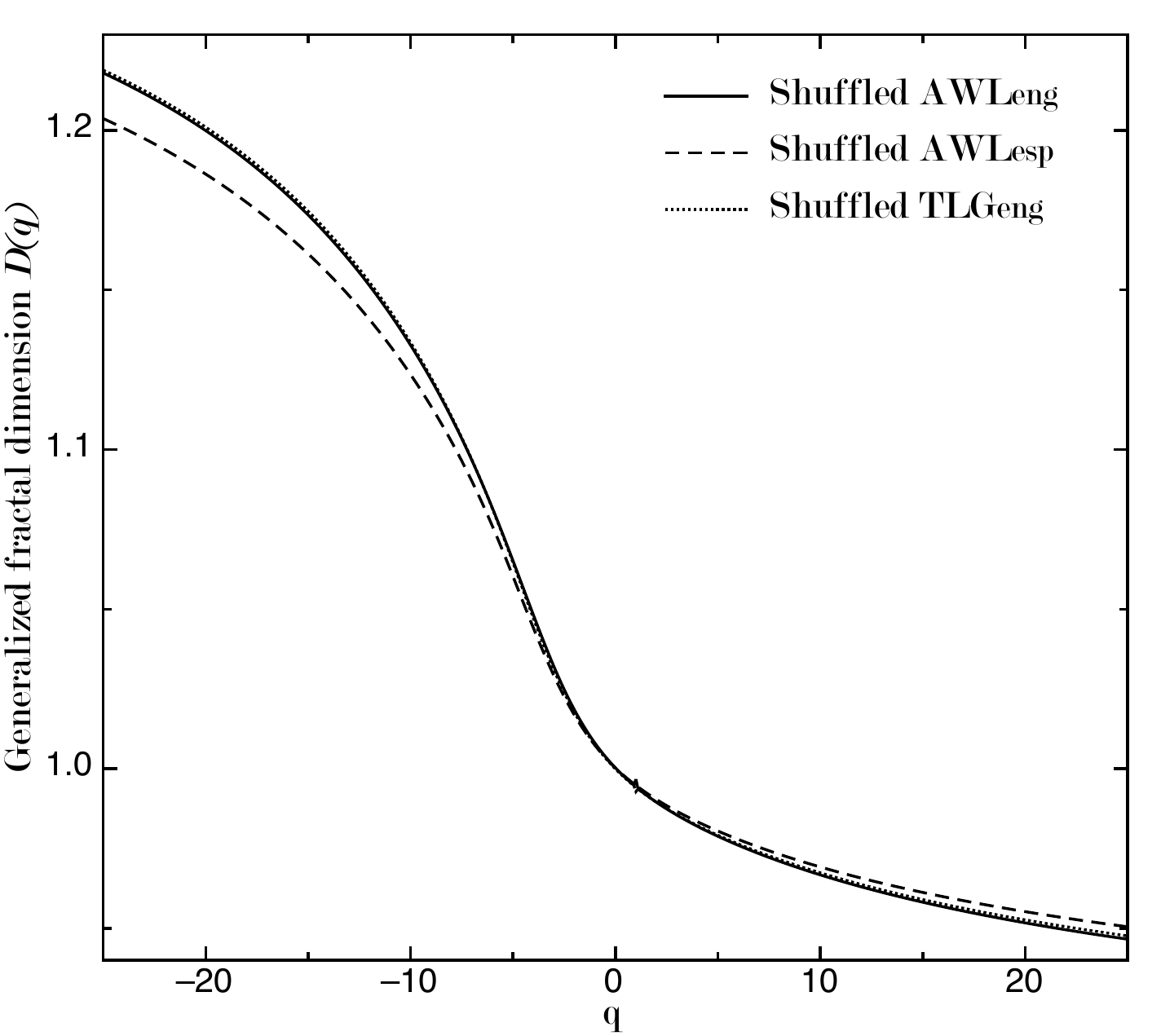}
\caption{\label{fig-2} D(q) for (a)  FTS (b) LTS  of the  shuffled indicated texts }
\end{figure}



\begin{figure}
\includegraphics[width=2.4in] {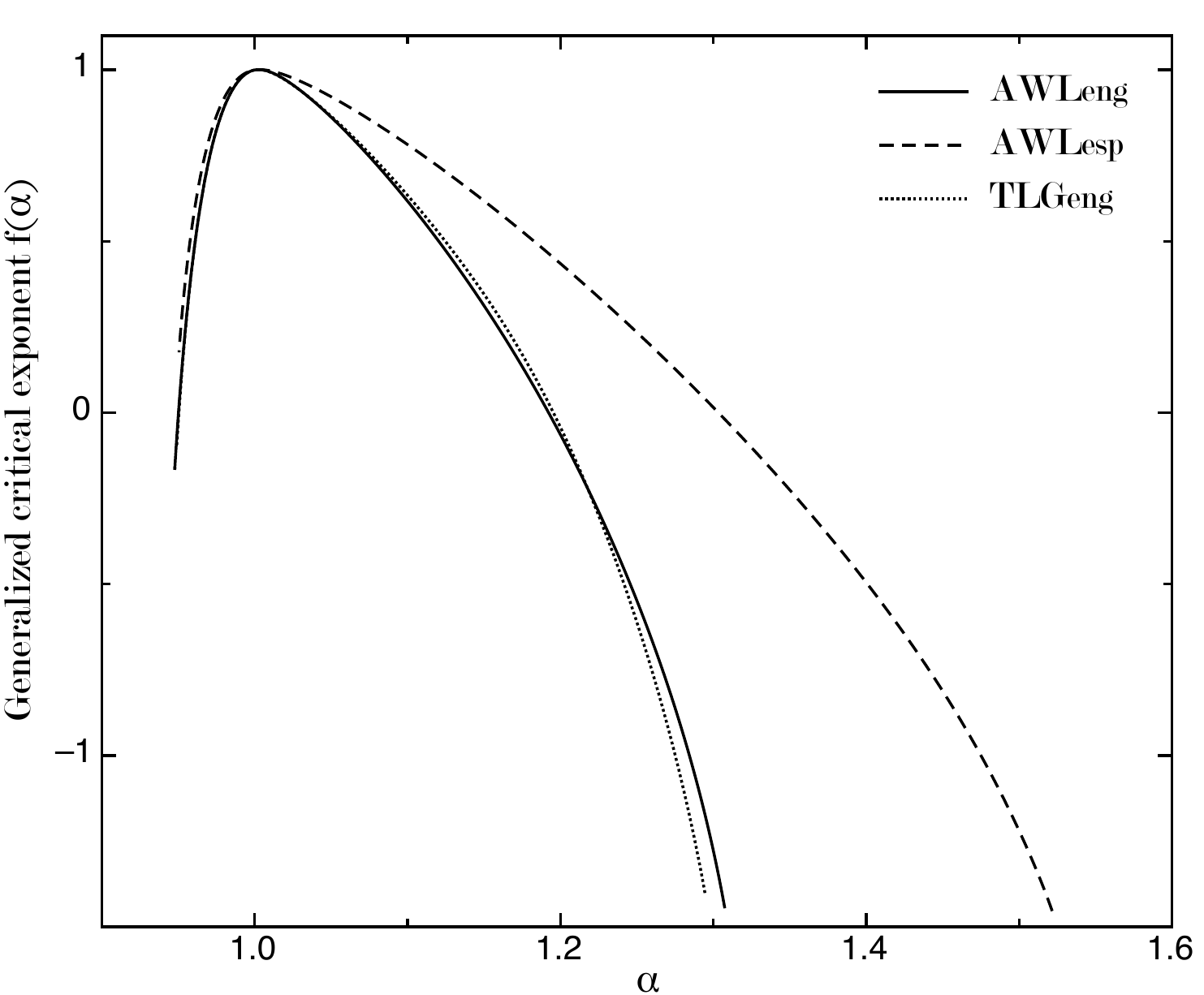}
\vspace{-0.4cm}
\includegraphics[width=2.4in] {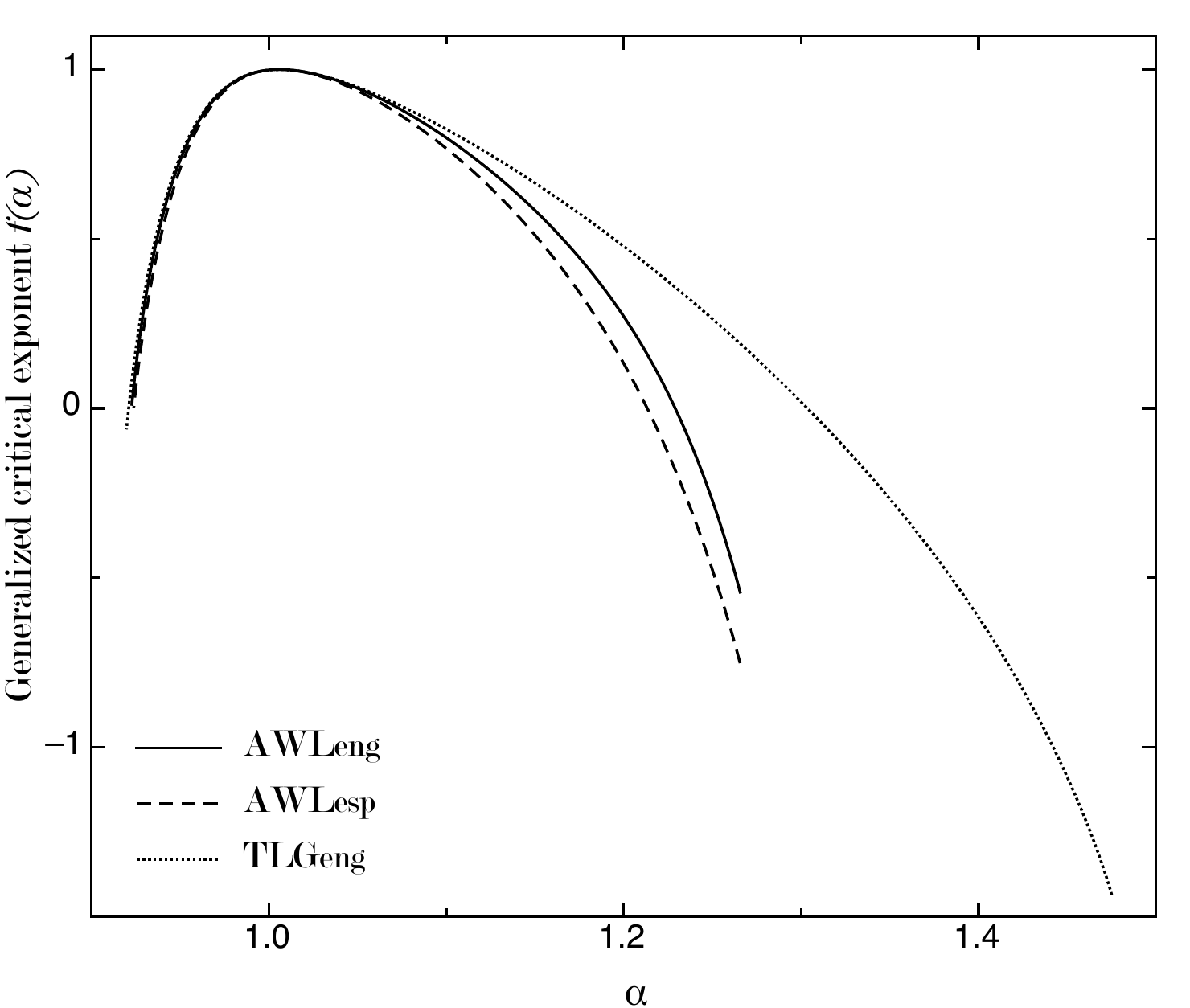}
\caption{\label{fig-5}   $f(\alpha)$ for (a)  FTS (b) LTS of the indicated texts}
\end{figure}

\begin{figure}
\includegraphics[width=2.4in] {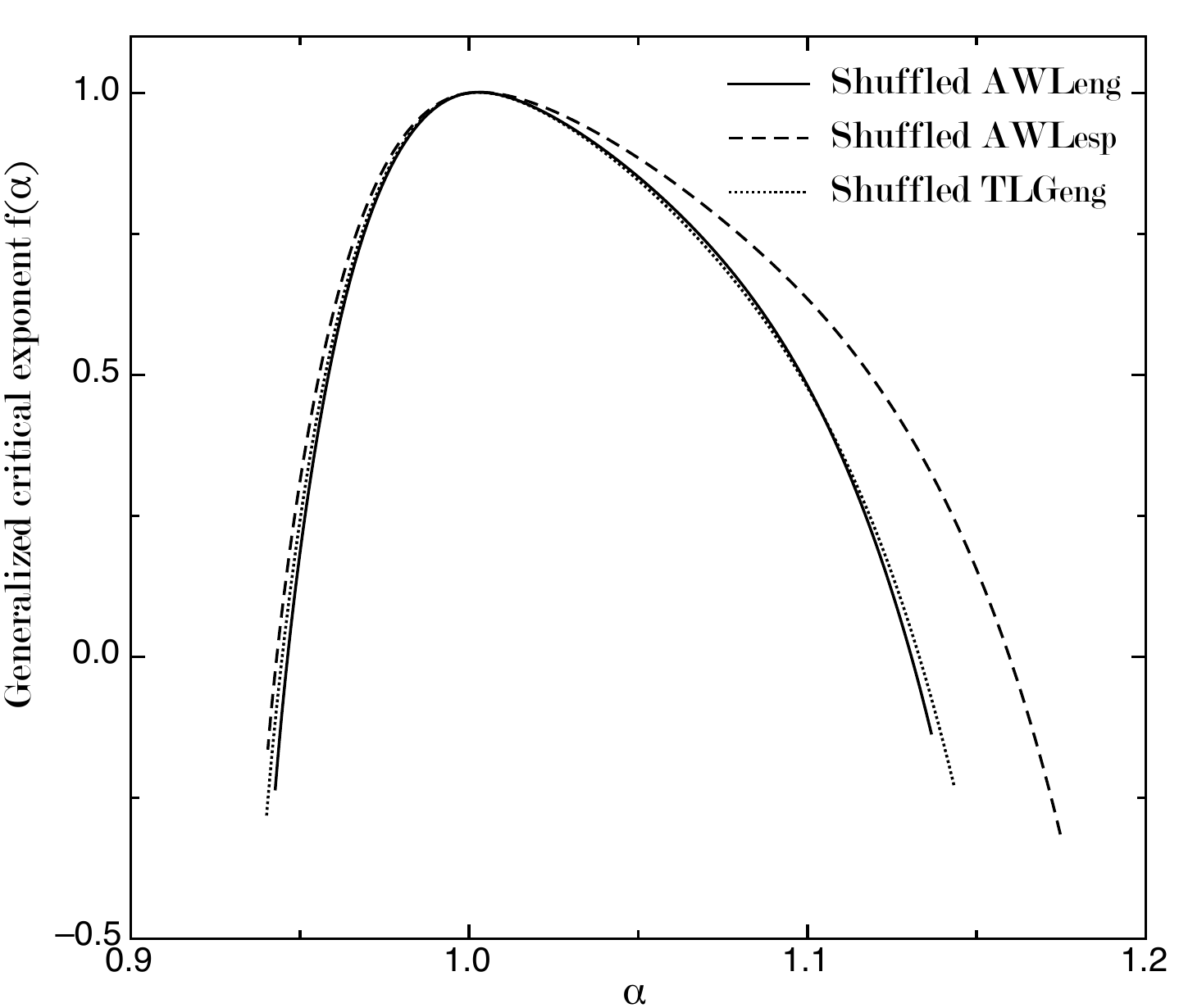}
\vspace{-0.4cm}
\includegraphics[width=2.4in] {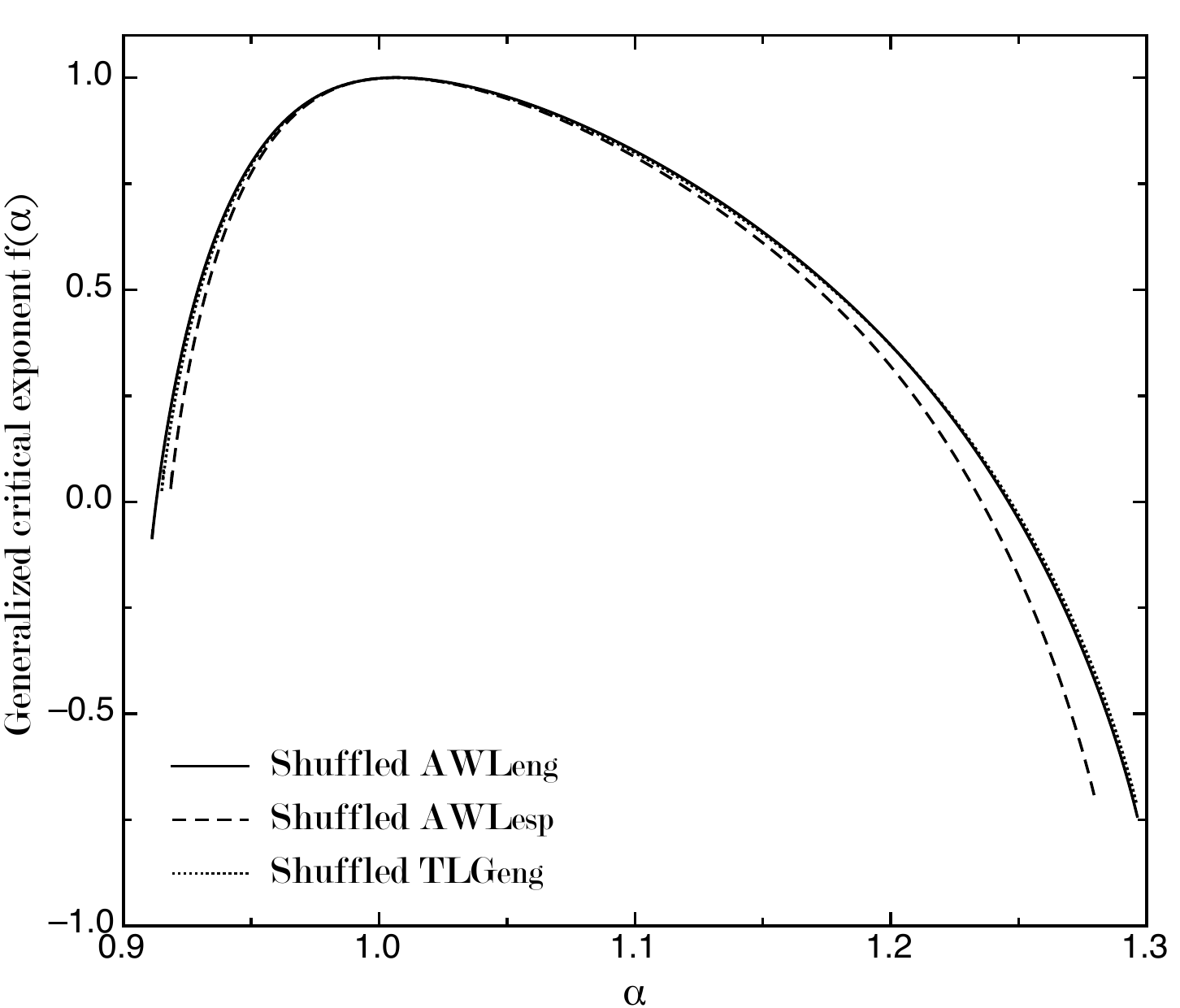}
\caption{\label{fig-6}  $f(\alpha)$ for (a)  FTS (b) LTS  of the  shuffled indicated texts}
\end{figure}

Obviously there is a large number of ways to map a text to a time series, but in the present study we only consider the above two since some physical meaning can be thought to arise in the mapping. 
As indicated in \cite{LTSpanos} the length of the word is associated with speaker effort, meaning that the longer the word the higher the effort required to pronounce it. The frequency of the word is also associated with the hearer effort as frequently used words require less effort to be understood by the hearer.

These time series are thereby analyzed along the multifractal ideas,  for which we briefly recall the formulae of interest in order to set the notations.

\subsection{Multifractal Analysis}

Let the (LTS ou FTS) time series having $N$ data points (words, here), i.e. $y_i$ ($1 \leq i \leq N$). 
 
Transform the series as follows: if the length of a word mot in LTS (or its  frequency in FTS) is smaller than the next one,  the former word gets a value = 2;  if it is greater, it gets   the value = 1; and 0 if both are equal.

The new series is called $ M_i$ ($1\leq i \leq N - 1 $). Each $ M_i $ is cut into $ N_s$ subseries of size $s$ , where $N_s$ is the smallest integer in $ N/s$. The ordering starts from the beginning of the text (contrary to analyses  in which some forecasting is expected, and for which the end ``points are more relevant).

Next one calculates the probability
\begin{equation}
P(s, \nu)  = \frac{\Sigma_{i=1}^{s}M_{(\nu-1)s+i}}{\Sigma_{\nu=1}^{N_s} \Sigma_{i=1}^{s}   M_{(\nu-1)s+i}}
\end{equation}
for every $\nu$ and $s$.  Thereafter one calculates
\begin{equation}
\chi(s, q) = \Sigma_{\nu=1}^{N_s}  P(s, \nu)^q
\end{equation}
for each $s$ value.  A power law behaviour  is expected
\begin{equation}
\chi(s, q) \sim  s^{\tau(q)};
\end{equation}
where $\tau(q)$ plays the role of the partition function \cite{halsey}. The generalized fractal dimension  $D(q)$ \cite{halsey,west} is defined from

\begin{equation}
D(q) = \frac{\tau(q)}{q-1}   
\end{equation}
and the generalized Hurst exponent, $h(q)$, from
\begin{equation}
h(q) = \frac{1+\tau(q)}{q} .
\end{equation}

Let 
\begin{equation}
\alpha = \frac{d\tau(q)}{dq}, 
\end{equation}
from which one obtains the generalized critical exponent, $f(\alpha)$ curve  \cite{12} from

\begin{equation}
f(\alpha) =  q \alpha - \tau(q).
\end{equation}

In the present work,  we have calculated $\chi(s,q)$  for $s$ between   2  and  200.  The $\tau(q)$ values were calculated by a linear best fit on a log-log plot of  $\chi(s,q)$ and $s$,  for $q$ values ranging from    -25 till 25.  
As one may expect the $q$ and $q-1$ values at the denominator in Eq.(4)-Eq.(5) were leading to numerical singularities. A smooth interpolation can be visually made without difficulty. For conciseness we don't show $\chi(s,q)$.


 \section{Results}  \label{sec-2}

The results of the FTS and LTS multifractal analysis for the three main texts  and their shuffled corresponding ones are shown in Figs. 1-4 (a-b). 

A mere perusing of the graphs indicate that the multifractal approach is in good order, e.g. since $D(q)$ is not a single point,  and should allow one to observe interesting LRO correlations and local ones.


\subsection{$D(q)$ plots: Figs. 1-2} 

In  FTS, the generalized fractal dimension has a similar set of values for both english texts, decaying from ca. 1.2 to 1.0 for $q$ increasing but negative; $D(q)$ decays slowly for $q$ positive, barely reaching a value 0.95  for $q$ = 80 (Fig. 1). The value of $D(q)$ is much  greater along the negative $q$ axis, for AWL$_{\mbox{esp}}$  but is identical to the other two for $q $  $\ge$ $0$. In LTS, even though the form of $D(q)$ is that to be expected, it has to be stressed that the  AWL$_{\mbox{eng}}$ and AWL$_{\mbox{esp}}$ are very quantitatively similar, but markedly differ from  TLG$_{\mbox{eng}}$. This already indicates that one can observe the high creativity of the author through these  two books. Moreover  the translation effect on style is much better seen on FTS than LTS.
 
The shuffled texts (Fig.2) remarkably have the same $D(q)$ values; their range and variations being similar to those of the real texts. Slight  quantitative differences occur, more markedly for the  shuffled AWL$_{\mbox{esp}}$ FTS, but  along a Baeysian reasoning these can be attributed to the finite size of the sample.

By the way,  
\begin{equation} C_1 =    \left.{d\tau(q) \over dq} \right|_{q=1} \end{equation}  
is a measure of the intermittency lying in the signal $y(n)$; it can be numerically estimated by measuring $\tau_q$ around $q=1$. In all cases the value of $C_1$ is close to unity.
Some conjecture on the role/meaning of $C_1$ is found  in Ref.\cite{4}.
 From some financial and political data analysis it seems that $H_1$ is a measure of the information entropy of the system. The same can be thought of here.

\subsection{$f(\alpha)$  plots: Figs. 3-4}
 
 The $f(\alpha)$  spectra are shown in Figs. 3-4. They are markedly  non symmetric, as was found for DLA \cite{DLA1,DLA2}, with very high positive skewness, i.e. for $q\le0$. Interestingly,  the esperanto text curve behaves differently from the english texts, in FTS, though TLG$_{\mbox{eng}}$ is different in the LTS case; the shuffled texts $f(\alpha)$ spectra behaving in a very similar  qualitative $and$ quantitative way. However the shuffling does not fully symmetrize the spectra.

It is interesting to observe that the $f(\alpha)$  curve is very sharp: it originates from negative values for $\alpha$ less than 1.0; reaches a maximum (=1.0) at 1.0, at the maximum so called box dimension, and decays rapidly for $\alpha$ positive;
  $f(\alpha)$ = 0 at $\alpha$=   1.2 and 1.3 respectively for AWL$_{\mbox{eng}}$ and TLG$_{\mbox{eng}}$; the maximum is also reached at (1.0, 1.0) and the spectrum spans the (narrow) interval  0.90: 1.25 for AWL$_{\mbox{esp}}$ on the $f(\alpha)$ =0 line. It is worth noticing that the values are reasonable in view of their correspondence to the fractal dimensions. On the other hand the sharpness indicates a high lack of uniformity of the texts LROC.

\section{Discussion with conclusion}  \label{sec-3}



In summary, one can observe similarities between the original and shuffled texts and their translations; see Table 1 for summarizing the similarities seen through $D(q)$ and $f(\alpha)$. The english  texts  look more similar with each other than with respect to the esperanto translation.
On the other hand, one physical conclusion arises from the above : the existence of a multifractal spectrum found for the examined texts indicates a multiplicative process  in the usual statistical sense for the distribution of words length and frequency in the text considered as a time series.  Thus linguistic signals may be considered indeed as the manifestation of a complex system of high dimensionality, different from random signals
or systems of low dimensionality such as the  financial and geophysical (climate) signals. 
Finally, the $f(\alpha)$ curve represents the measurable aspects of the word networks, be they considered through LTS or FTS. Our work confirms that texts could be seen as networks indeed \cite{rodgers}.

Before suggesting a physical model describing the construction of a writing, let us consider implications from the  $f(\alpha)$ spectrum in some detail.
The not fully parabolic, to say the least, $f(\alpha)$ curve indicates non uniformity and strong LROC between long words and small words, - evidently arising from strong short range correlations between these. In some sense this is expected for classical writings.  It is usually known that the left  (right) hand side of the  $f(\alpha)$ curve  corresponds to fluctuations of the $q\ge0$ ($q\le0$)-correlation function. In other words, they correspond to fluctuations in small (large) word distributions. Therefore the distribution of small and long words should be examined in further work in order to observe these local correlations, e.g. through a detrended fluctuation analysis. 

\begin{table}
\begin{footnotesize}
\begin{center} 
\begin{tabular}{|c|cc|cc|  } \hline 
Series & $D(q)$ & $D^s(q)$	&  $ f(\alpha)$ &  $f^s(\alpha)$ \\ \hline 
&  AWL$_{\mbox{eng}}$ & AWL$_{\mbox{eng}}$ 	&AWL$_{\mbox{eng}}$  & AWL$_{\mbox{eng}}$ \\
FTS &  $ \simeq$  & $ \simeq$ &$ \simeq$  & $ \simeq$   \\
&	TLG$_{\mbox{eng}}$ &  $TLG_{\mbox{eng}}$  &$TLG_{\mbox{eng}}$ &  $TLG_{\mbox{eng}}$   \\ \hline   
&	AWL$_{\mbox{eng}}$ &  AWL$_{\mbox{eng}}$ &AWL$_{\mbox{eng}}$  &  AWL$_{\mbox{eng}}$   \\ 
LTS&	$ \simeq$  &$ \simeq$   & $ \simeq$  &	$ \simeq$ 	 \\
&	AWL$_{\mbox{esp}}$ &  $TLG_{\mbox{eng}}$   &AWL$_{\mbox{esp}}$&  $TLG_{\mbox{eng}}$  \\ \hline
\end{tabular}
\label{tab01similarit} 
\end{center}
\end{footnotesize}
\caption{Comparing original texts  quasi identical behaviors through functions $D(q)$ and $ f(\alpha)$, and their counterpart for shuffled ($s$) cases, i.e. $D^s(q)$ and $ f^s(\alpha)$}
\label{tab-2}
\end{table}

\begin{table}
\begin{footnotesize}
\begin{center}  
\begin{tabular}{|c| c c c c c c c| c c c c c c c|} \hline 
\centering{Original  texts} &  Q	& $\alpha_-$& $\alpha_+$ & $w_1$ &$w_2$& $r_1$&$r_2$ \\ \hline 
AWL$_{\mbox{eng}}$ FTS &    5.71 &0.95&1.19&0.96&0.04&0.97&0.03 \\ 
AWL$_{\mbox{esp}}$ FTS &    4.39 &0.94&1.30&0.99&0.01&0.99&0.01	\\ 
TLG$_{\mbox{eng}}$  FTS &    5.71 &0.95&1.19&0.99&0.01&0.99&0.01 \\ \hline
AWL$_{\mbox{eng}}$ LTS &    4.65&0.92&1.23&0.89&0.11&0.91&0.09	 \\ 
AWL$_{\mbox{esp}}$ LTS &    4.83 &0.92&1.21&0.87&0.13&0.89&0.11	\\ 
TLG$_{\mbox{eng}}$  LTS &   3.94 &0.92&1.34&0.96&0.04&0.97&0.03 \\ \hline \hline
\centering{Shuffled  texts} &  Q	& $\alpha_-$& $\alpha_+$ & $w_1$ &$w_2$& $r_1$&$r_2$ \\ \hline 
AWL$_{\mbox{eng}}$ FTS& 6.94 &0.95&1.13&0.89&0.11&0.90&0.10 \\ 
AWL$_{\mbox{esp}}$ FTS& 6.57 &0.96&1.16&0.97&0.03&0.97&0.03 \\ 
TLG$_{\mbox{eng}}$  FTS& 6.59 &0.94&1.13&0.82&0.18&0.84&0.16	 \\ \hline 
AWL$_{\mbox{eng}}$ LTS& 4.35 &0.91&1.25&0.88&0.12&0.90&0.10 \\ 
AWL$_{\mbox{esp}}$ LTS& 4.56 &0.92&1.24&0.90&0.10&0.92&0.08 \\ 
TLG$_{\mbox{eng}}$  LTS& 4.35 &0.91&1.25&0.88&0.12&0.9&0.10	 \\ \hline 
\end{tabular}
\end{center}
\end{footnotesize}
\caption{ Tsallis $Q$-parameter, derived from $\alpha_-$ and $\alpha_+$ values, read on Figs. 3-4, from Eq. (11),  for the original texts and for shuffled or translated corresponding texts, according to the type of series;  the weights and ratios of the binomial cascade approximation (see text) are given}
\label{tab-3}
\end{table}

Moreover, in order to characterize the writings, texts and/or authors,  we propose a very rough approximation/model, i.e. let us consider (assume !) that the writings are made of only two types of  words : small and large \cite{footnote3,blogs}, appearing through some recursive process. In so doing one can consider the behavior of the atypical $f(\alpha)$ curve as originating form a binomial cascade of short and long words, on a support [0,1], with an arbitrary contraction ratio $r_i$ and a weight $w_i$ for the word in each successive subinterval, as for a  multifractal Cantor set construction \cite{halsey}. For an arbitrary number $n$ of subintervals the generalized fractal dimension (or rather $\tau(q)$) is obtained from
\begin{equation}
\sum_{i=1}^{n} w_i^q r_i^{-\tau}= 1.
\label{generalwr}
\end{equation}
The formula is easily generalized for random contraction ratios and weights. However it simplifies for the case of a simple binomial cascade, i.e.
\begin{equation}
 w_1^q r_1^{-\tau} +  w_2^q r_2^{-\tau} =1.
\label{binomwr}
\end{equation}
Whence  the extremal $\alpha$ values read $\alpha_- = \log (w_2)/ \log(r_2)$ and $\alpha_+=\log (w_1)/\log(r_1)$, from which the weights and ratios can be estimated by inversion (table \ref{tab-2}), thereby suggesting the author's somewhat systemic way used in his/her writings.


The physics connection can be obtained if one relates the $f(\alpha) $ curve extremal points through their physical meanings \cite{arimitsu}, i.e.

 \begin{equation}
\frac{1}{1-Q } = \frac{1}{\alpha_+} - \frac{1}{\alpha_-}
\label{eqQ}
\end{equation}

 where $\alpha_-$ and $\alpha_+$ are the extremes of the range of support for the (positive) multifractal spectrum $f(\alpha)$ and $Q$ (instead of the usual $q$) is used to represent the  parameter arising in the non extensive description of statistical physics \cite{tsallis1}; see values in table \ref{tab-2}. By extension it is a measure of  the attractor dimension or the number of so called degrees of freedom. It is obvious from the table that Q varies between 4 and 7, with interesting differences between the LTS and FTS cases, LTS's Q being systematically smaller, in the original or shuffled texts. Notice that the value of Q is   more extreme though with the same order of magnitude in the case of the esperanto text for both types of series.

Finally we re-emphasize the remarkable difference for the esperanto text (Fig. 3a) with the english texts in the FTS analysis. Linguistics input should be searched at this level and is left for further discussion.  The origin of differences between TLG and AWL needs more work at the linguistic level. However we have indicated the interest of the multifractal scheme  in providing a measure of these correlations, thus a new measure of an author's style. This suggests  a (binomial, at first) cascade model containing parameters characterizing (or reflecting, at least) the text style, and most likely {\it in fine} the author writings.  It remains to be seen whether the $f(\alpha)$ curve and the  (to be generalized) binomial cascade model, with the weight and ratio parameters hold through in other cases, and can characterize authors and texts, - and in general time series.  Moreover the multifractal  method should additionally be able to distinguish a natural
language signal from a computer code signal \cite{LTSpanos} and help in improving translations by suggesting perfection criteria and indicators of text qualitative values.

{\bf Acknowledgements}
The authors would like to thank D. Stauffer for as usual fruitful discussions ....
This work  has been supported by European Commission Project 
CREEN FP6-2003-NEST-Path-012864.


\begin{thebibliography}{0}
\bibitem{footnote1} When finalizing the writing of this paper we became aware of related work of interest  \cite{newIJMPC}.
\bibitem{footnote2} The shuffle algorithm is found on wikipedia. The first ``data point is exchanged with a following  one, its location chosen from a generated random number. The second data point is exchanged with  a  following one, chosen from another random number, etc.  The random number generator was check to lead to a rather uniform distribution, for a number between 0 and 1. The algorithm was applied ten times on the texts to get the final shuffled texts we used.
\bibitem{footnote3} We apologize to all authors for this simplification of their  work.  Yet, in some sense this is equivalent to considering that the (+/-1) Ising model describes all magnetic features. Notice that this assumption on writings has been successfully made in the study of blogs, \cite{blogs} where words occurring with an equivalent frequency are considered to be $identical$ for studying the blogs statistical properties.
\end{thebibliography}
\end{document}